%% file: main.tex
\documentclass[10pt,twocolumn,letterpaper]{article}

\usepackage[pagenumbers]{cvpr} 

\input{preamble}

\definecolor{cvprblue}{rgb}{0.21,0.49,0.74}
\usepackage[pagebackref,breaklinks,colorlinks,allcolors=cvprblue]{hyperref}
\usepackage{multirow}
\usepackage{makecell}
\usepackage{hypcap}

\usepackage{listings}
\usepackage{xcolor}
\lstdefinelanguage{json}{
    basicstyle=\ttfamily, 
    showstringspaces=false,
    commentstyle=\color{gray}, 
    stringstyle=\color{brown}, 
    keywordstyle=\color{blue}, 
    morestring=[b]",
    morekeywords={true,false,null} 
}
\def\paperID{5247} 
\def\confName{CVPR}
\def\confYear{2025}

\author{Zejian Li$^1$
\and Chenye Meng$^1$ \and Yize Li$^2$ \and Ling Yang$^3$ \and Shengyuan Zhang$^1$ \and Jiarui Ma$^1$ \and Jiayi Li$^2$ \and Guang Yang$^4$ \and Changyuan Yang$^4$ \and Zhiyuan Yang$^4$ \and Jinxiong Chang$^5$ \qquad Lingyun Sun$^1$ \\
{\small $^1$ Zhejiang University \quad \small $^2$ Jiangnan University \quad \small $^3$ Peking University \quad \small $^4$ Alibaba Group \quad \small $^5$ Ant Group} \\
{\tt\small $^1$ \{zejianlee,mengcy\}@zju.edu.cn} \\
}

\begin{document}
\title{LAION-SG: An Enhanced Large-Scale Dataset for Training Complex Image-Text Models with Structural Annotations}

\maketitle
\input{sec/mcy_main}

{
    \small
    \bibliographystyle{ieeenat_fullname}
    \bibliography{main}
}
\newpage 


\begin{center}
    \section*{Supplementary Material}
\end{center}
\input{sec/mcy_SM}

\end{document}

%% file: preamble.tex
\usepackage{ulem}

%% file: sec/mcy_main.tex
\begin{abstract}
Recent advances in text-to-image (T2I) generation have shown remarkable success in producing high-quality images from text. 
However, existing T2I models show decayed performance in compositional image generation involving multiple objects and intricate relationships.
We attribute this problem to limitations in existing datasets of image-text pairs, which lack precise inter-object relationship annotations with prompts only. 
To address this problem, we construct LAION-SG, a large-scale dataset with high-quality structural annotations of scene graphs (SG), which precisely describe attributes and relationships of multiple objects, effectively representing the semantic structure in complex scenes.
Based on LAION-SG, we train a new foundation model SDXL-SG to incorporate structural annotation information into the generation process. 
Extensive experiments show advanced models trained on our LAION-SG boast significant performance improvements in complex scene generation over models on existing datasets. 
We also introduce CompSG-Bench, a benchmark that evaluates models on compositional image generation, establishing a new standard for this domain. 
Our annotations with the associated processing code, the foundation model and the benchmark protocol are publicly available at \href{https://github.com/mengcye/LAION-SG}{https://github.com/mengcye/LAION-SG}.
\end{abstract}

\section{Introduction}
\label{sec:intro}

Text-to-image (T2I) generative models~\cite{saharia2024Photorealistic_imagen, dalle2, DM_beats_GAN, Chen2024PixArtWT, Tewel2024TrainingFreeCT, Shifted_Diffusion, Generative_Image_Dynamics, podell2023sdxlimprovinglatentdiffusion} have made significant strides, showcasing an impressive ability to generate high-quality images from text prompts. However, when reviewing T2I generation models, it is observed that they are generally effective for simple scene generation but notably deteriorate when handling complex scenes—such as those involving multiple objects and intricate relations between them (~\cref{fig: teaser}). We attribute this limitation to a lack of emphasis on complex inter-object associations within existing text-image datasets. Prior T2I works have primarily focused on architectural improvements, which fail to address this underlying issue.

Scene graphs (SG) provide a structured description of image content. A scene graph consists of nodes (representing objects and attributes) and edges (depicting relationships between objects). Compared to the sequential description of text, SGs offer compact, structured approaches describing complex scenes, enhancing annotation efficiency. SGs also allow for more precise specification of specific objects associated attributes and their relationships, a feature critical for generating complex scenes. However, existing scene graph datasets are relatively small in scale (e.g., COCO-Stuff~\cite{caesar2018coco} and Visual Genome~\cite{krishna2017visual}), while large-scale datasets primarily consist of text annotations only.

Our work focuses on compositional image generation via scene graphs (SG2CIM). 
We construct LAION-SG dataset, a significant extension of LAION-Aestheics V2 (6.5+)~\cite{LAION_5B} with high-quality, high-complexity scene graph annotations. Our annotations feature multiple objects, attributes and relationships describing images of high visual quality. 
Therefore, our LAION-SG better encapsulates the semantic structure of complex scenes, supporting improved generation for intricate scenarios. 
The advantage of LAION-SG in complex scene generation is validated in further experiments with multiple metrics on semantic consistency.

With LAION-SG, we train existing models and propose a new baseline for generating complex scenes with SGs. 
To build the baseline, we use SDXL~\cite{podell2023sdxlimprovinglatentdiffusion} as the backbone model and train an auxiliary SG encoder to incorporate SG within the image generation process.
Specifically, the SG encoder employs a graph neural network (GNN)~\cite{scarselli2008graph_GNN} to extract scene structure in graphs, thereby optimizing the SG embeddings. These embeddings are then fed into the backbone model to yield high-quality complex images. 
Our approach efficiently enhances the model’s ability to generate complex scenes and provides a foundational model.

Finally, we establish CompSGen Bench, a benchmark for complex scene generation evaluation. Using this benchmark, we evaluate existing state-of-the-art (SOTA) models alongside our baseline variants trained on COCO-Stuff~\cite{caesar2018coco}, Visual Genome~\cite{krishna2017visual}, and our LAION-SG. Both quantitative and qualitative results demonstrate models trained on LAION-SG outperforming other model trained on COCO-Stuff and Visual Genome as in current T2I and SG2IM baselines. We conclude that LAION-SG dataset significantly enhances complex scene generation. Our LAION-SG dataset represents a pioneering effort in annotating complexity on existing image datasets and holds potential for wider applications for scene perception and synthesis.

In summary, our contributions are as follows:
\begin{itemize}
    \item We introduce a new scene graph (SG)-based dataset, LAION-SG, for complex scene image generation. This dataset includes high-quality SG annotations with multiple objects, their attributes, and numerous relationships, enhancing generative models' capability to handle complex scenes and improving the complexity and fidelity of generated images.
    \item We fine-tune a new efficient foundation model on the proposed dataset. The model demonstrates heightened sensitivity to image content awareness and competitive performance in complex scene generation, setting a new baseline for SG-image understanding.
    \item We establish CompSGen Bench, a benchmark for complex scene generation evaluation with several metrics, and conduct extensive experiments to validate the effectiveness of our dataset and baseline.
\end{itemize}

\section{Related Work}
\textbf{Compositional Image Generation.} 
Text-to-image generation~\cite{saharia2024Photorealistic_imagen, dalle2, DM_beats_GAN, Chen2024PixArtWT, Tewel2024TrainingFreeCT, Shifted_Diffusion, Generative_Image_Dynamics, podell2023sdxlimprovinglatentdiffusion} has advanced significantly, particularly through diffusion models~\cite{ddpm, ldm}.
However, the sequential format of the textual data imposes limitations on image generation. This limitation is particularly evident when generating compositional images that involve multiple objects and associated relationships \cite{yang2024mastering,lian2023llmgrounded,zhang2024itercomp}. 


Previous studies have explored various methods to enhance the controllability of the text-to-image diffusion model.
Compositional Diffusion~\cite{Compositional_Diffusion} breaks down complex text prompts into multiple easily generated segments, but it is limited to conjunction and negation operators. Attend-and-Excite~\cite{Attend-and-Excite} guides pre-trained diffusion models to generate all entities in the text through immediate reinforcement activation, yet it still faces attribute leakage issues.

Other approaches use additional spatial conditions to improve generation. GLIGEN~\cite{Li2023gligen} integrates bounding boxes by adding trainable gated self-attention layers while freezing the original weights. Ranni~\cite{Feng2024Ranni} incorporates a semantic panel containing layouts, colors, and keypoints parsed from Large Language Model (LLM) for training diffusion models. However, these efforts require costly training. 

Universal Guidance~\cite{Bansal2023Universal} utilizes a well-trained object detector and constructs new losses to enforce the generated images to match the positional guidance. BoxDiff~\cite{Xie2023BoxDiff} encourages the desired objects to appear in specified regions by calculating losses based on the maximum values in the cross-attention maps. RealCompo~\cite{zhang2024RealCompo} achieves a balance between the realism and complexity of images by adjusting the influence of textual and layout guidance in the denoising process. Nevertheless, these methods rely on the accuracy of the initial bounding boxes, despite some works~\cite{Feng2024LayoutGPT, lian2023llmgrounded, zhang2023controllable} attempt to directly generate spatial layout information based on textual conditions using LLM.

For text-to-image generation, all of these methods mainly focus on model improvement, failing fundamentally to address the limitations imposed from the dataset.

\textbf{Image Generation from Scene Graphs} 
(SG2IM)~\cite{johnson2018image, krishna2017visual} involves creating images based on structured representations of scenes, where objects and their relationships are explicitly defined as a graph~\cite{xu2017scene}.

The traditional SG2IM methods typically consist of two stages~\cite{johnson2018image, ashual2019specifying, Compositional_Diffusion, Du2023Reduce}. They first transform the input scene graph into a layout, which serves as an auxiliary for image refinement using a generative model. 
These methods are effective for generating images with a few objects and simple relationships, 
but they often generate confusingly when relations become abstract and the number of objects increases.

Therefore, some methods propose directly learning the alignment from scene graphs to images~\cite{feng2023trainingfree, Wang2024compositional, Wu2023Scene, LOCI}. SGDiff~\cite{Yang2022DiffusionBasedSG} pre-trains an SG encoder using contrastive learning, which is combined with Stable Diffusion (SD) to generate images. SG-Adapter~\cite{Shen2024SGAdapterET} is trained to fine tune SD, incorporating SG information into the text-to-image process through attention layers. R3CD~\cite{Liu2024R3CD} enhances large-scale diffusion models by leveraging SG transformers to learn abstract interactions and improve image generation.
These methods overcome the issues associated with sequential text conditions, enhancing the model's semantic expressive capacity. 
However, the current SG-image datasets do not match the scale and quality of text-image datasets, resulting in a quality bottleneck for SG2IM methods.

\begin{figure*}[t]
    \centering
    \includegraphics[width=1\textwidth]{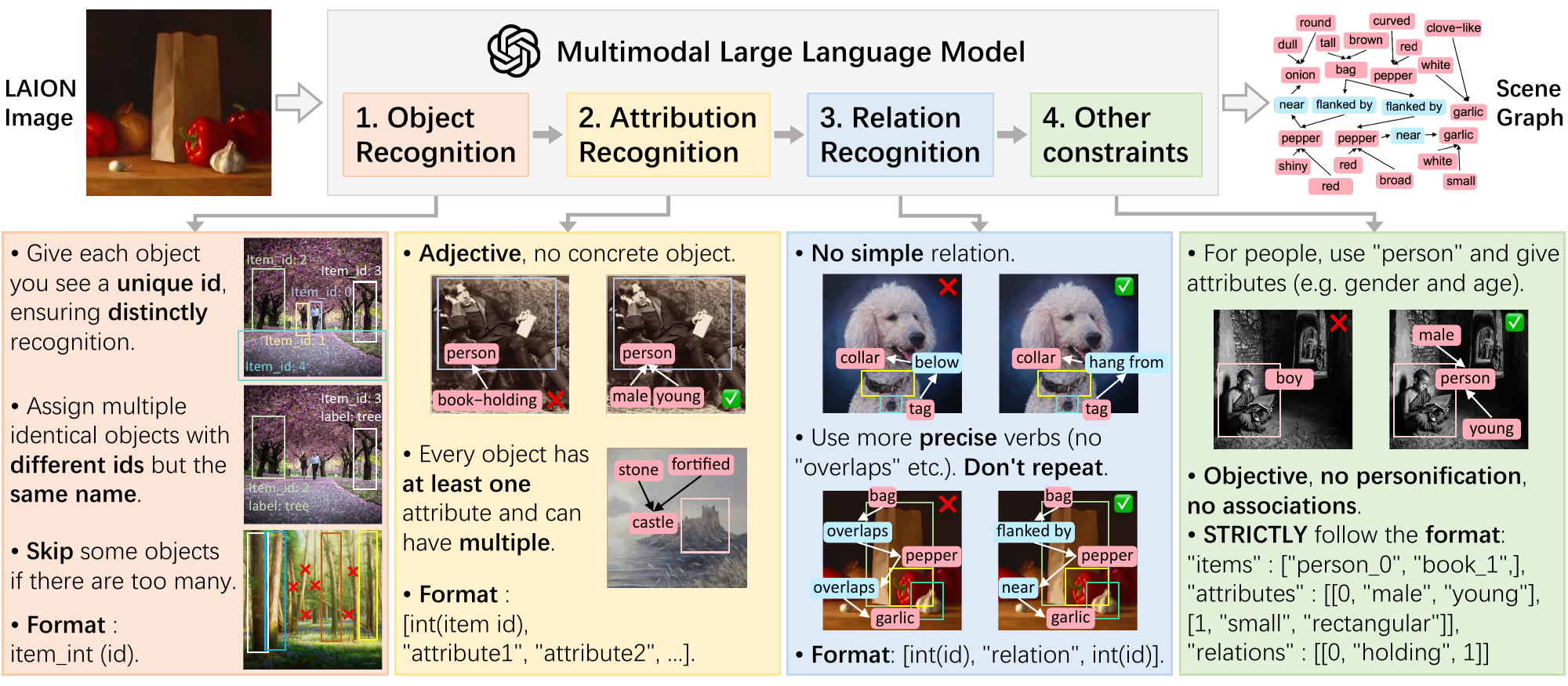} 
    \vspace{-1.5em}
    \caption{The construction pipeline of LAION-SG dataset. 
    1) Identify the objects in the image and assign a unique ID to each. 2) The attributes must be abstract adjectives and should not include specific objects. Each object may have one or more attributes. 3) The relations between objects should be as specific as possible, avoiding simple relations. Use more precise verbs, minimizing repetition. 4) For people, label the object as ``person'' and include attributes such as gender and age. Avoid anthropomorphism or associations, and provide an objective description of what is observed in the image.}
    \label{fig2}
    \vspace{-1em}
\end{figure*}

\textbf{Large-Scale Image-Text Datasets and Benchmarks.}  
Previous datasets, such as MS-COCO~\cite{Lin2014MicrosoftCC}, Visual Genome~\cite{krishna2017visual}, and ImageNet~\cite{ImageNet}, are primarily constructed through manual annotation. These datasets are highly accurate in terms of labeling but limited in scale due to the considerable costs associated with manual annotation. To address the limitations, some approaches propose automatically extracting and filtering data from websites. The Conceptual Captions dataset (CC3M)~\cite{Sharma2018ConceptualCA_CC3M}, for instance, contains approximately 3.3 million image-text pairs, and its extended version, CC12M~\cite{Changpinyo2021Conceptual1P_CC12M}, increases the dataset size to 12 million pairs. LAION-5B~\cite{LAION_5B} further expands dataset scale, comprising around 5.85 billion image-text pairs, making it one of the largest publicly available image-text datasets. 
LAION-Aesthetics, a subset of LAION-5B, is curated for high visual quality and intended to support image generation and aesthetic research. However, 
it does not always ensure textual descriptions that accurately reflect image content. 
Thus we enhance LAION-Aesthetics with structured annotations in the form of Scene Graphs (SG), 
constructing a high-quality, large-scale dataset tailored for generating compositional images.

In text-to-image generation, several benchmarks comprehensively assess model performance across various aspects. T2I-CompBench~\cite{T2I_CompBench} provides 6,000 compositional prompts covering attribute binding, object relationships, and complex compositions. HRS-Bench evaluates models on 13 skills across 50 scenarios, including accuracy, robustness, and bias~\cite{Bakr_2023_ICCV_HRS_Bench}. HEIM assesses 12 dimensions, such as text-image alignment, aesthetics, and multilinguality~\cite{NEURIPS2023_dd83eada_HEIM}. VISOR focuses on spatial relationships with its $\mathrm{SR}_{2D}$ dataset~\cite{gokhale2023benchmarkingspatial_VISOR}, and HPS v2 enables comparison based on human preferences~\cite{HPSv2}. 
These benchmarks only focus on text-based image generation. 
To fill the gap in this domain, we are the first to propose a complex scene generation benchmark based on scene graphs.
%



\section{Dataset and Benchmark}
A large-scale, high-quality dataset is essential for learning compositional image generation. However, existing large-scale T2I datasets, such as LAION, describe content beyond the images (as illustrated in ~\cref{fig: qualitative}), 
misleading the generation. In contrast, SG datasets tend to focus more specifically on the actual content within images, namely the objects and relationships. Nonetheless, current SG datasets, such as COCO and VG, are relatively small in scale and have limited object and relationship types, making them insufficient for compositional image generation.

To address this, we propose LAION-SG dataset, which is a large-scale, high-quality, open-vocabulary SG dataset. We also introduce the Complex Scene Generation Benchmark (CompSGen Bench) to evaluate models’ performance . 

\begin{figure*}[t]
    \centering
    \includegraphics[width=1\textwidth]{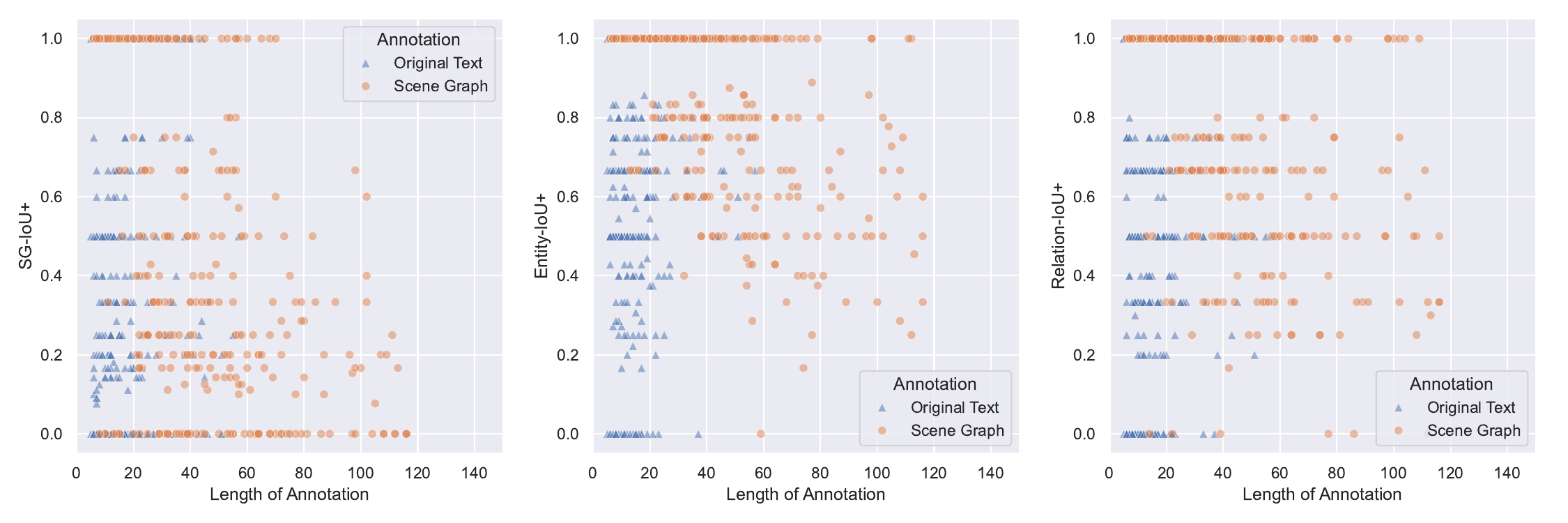} 
    \vspace{-2em}
    \caption{The annotation length and accuracy characteristics of LAION-SG compared to the LAION-Aesthetics. Compared to text, the scene graph, as a more compact form, has a longer length and its accuracy is more concentrated in high-scoring areas. 
    This suggests that our LAION-SG annotation more accurately reflects the image information and contains richer semantics.
    }
    \label{fig: scatter}
    \vspace{-1.5em}
\end{figure*}

\begin{figure*}[!ht]
    \centering
    \includegraphics[width=1\textwidth]{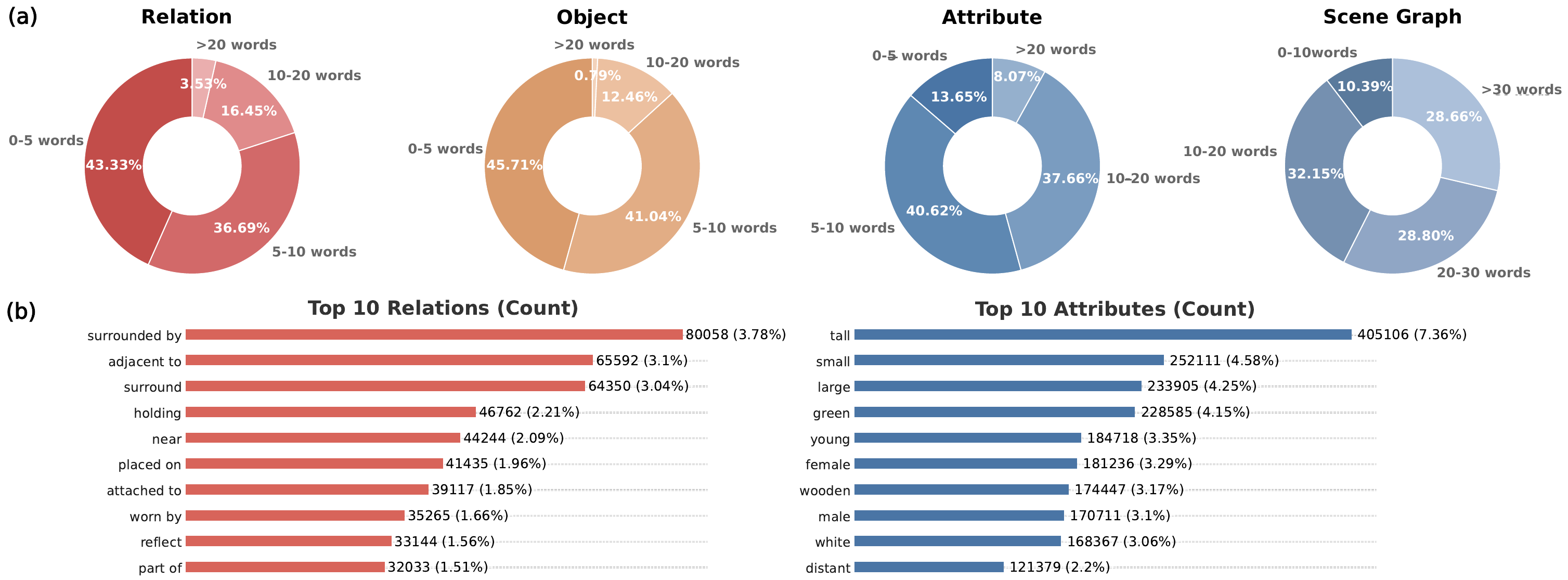} 
    \vspace{-1.5em}
    \caption{The annotation distribution of LAION-SG. (a) The length of the scene graph lies in a wide range. Our annotation provides more specific information compared to single-word descriptions, while also avoiding the inefficiency in model learning caused by excessively lengthy annotations. (b) The top 10 relations and attributes represent only a small percentage of the total distribution, indicating that LAION-SG covers a highly diverse range of annotations, showcasing its large scale and open vocabulary.
    }
    \label{fig:bar_donut}
\end{figure*}

\begin{table*}
  \centering
  \begin{tabular}{@{}lccccc@{}}
    \toprule
    \textbf{Annotation} & \# \textbf{Objects (w/o Proper Noun)} & \textbf{Length} & \textbf{SG-IoU+}$^\uparrow$ & \textbf{Ent.-IoU+}$^\uparrow$ & \textbf{Rel.-IoU+}$^\uparrow$ \\
    \midrule
    LAION Caption & 5.33$\pm$3.94 (2.02$\pm$3.01) & 19.0$\pm$19.7 & 0.306 & 0.631 & 0.557\\
    LAION-SG (Ours) & 6.39$\pm$4.17 & 32.2$\pm$20.3 & \textbf{0.422} & \textbf{0.810} & \textbf{0.749} \\
    \bottomrule
  \end{tabular}
  \vspace{-0.5em}
  \caption{The number of objects and length per sample, and the average accuracy for 300 samples across different annotation types.}
  \label{tab: statistic feature of dataset}
  \vspace{-1em}
\end{table*}

\subsection{Dataset Construction}
Our LAION-SG dataset is built on high-quality images in LAION-Aesthetic V2 (6.5+)~\cite{LAION_5B} with automated annotation performed using GPT-4o~\cite{openai2024gpt4technicalreport}. LAION-Aesthetics V2 (6.5+) is a subset of LAION-5B~\cite{LAION_5B}, comprising 625,000 image-text pairs with predicted aesthetic scores over 6.5, curated using the LAION-Aesthetics Predictor V2 model. 
During our construction, only 540,005 images are available in the dataset due to copyright or other issues.

Through prompt engineering, we devised a set of specific requirements for scene graph annotations to ensure comprehensiveness, systematic structure, and precision in the annotation results. \cref{fig2} illustrates the detailed construction pipeline of LAION-SG. 
Each component plays a crucial role in achieving high-quality automated annotation.

First, as scene graphs typically contain multiple objects and their relations, the prompt requires 
``identification of as many objects, attributes, and their relations within the image as possible''. 
This design encourages that all objects and interactions in a scene are annotated.
Each object is assigned a unique ID, even for multiple objects of the same type, ensuring that the entirety of the scene's structure and hierarchy is accurately represented.

Second, the attribute section mandates that each object must have at least one abstract adjective attribute, while avoiding the use of other objects as attributes. This design is especially important in complex scenes as it helps differentiate objects' appearance, state, and characteristics from the background and other elements, maintaining consistency and clarity in annotations. 
By avoiding the confusion between specific objects and abstract attributes, the annotations become more interpretable and generalizable.

In the relation section, we specify the use of concrete verbs to describe relations between objects rather than relying solely on spatial orientation. 
This is because relations are often more critical in scene graphs than mere spatial information. 
By using precise verbs like ``standing on'' or ``holding'', we capture dynamic interactions within the scene, which is essential for complex scene generation.

Leveraging these prompts with the multimodal large language model GPT-4o, we generate annotations representing scene graphs.
Our annotation is expect to achieve accuracy for every object, attribute, and relationship, thoroughly covering each detail in the scene and providing robust data support for subsequent compositional image generation tasks.

\subsection{LAION-SG Dataset}
\label{subsec: LAION-SG dataset}
By performing the construction strategy, we develop LAION-SG, a large-scale, high-quality dataset containing 540,005 SG-image pairs annotated with objects, attributes, and relationships. This dataset is divided into a training set of 480,005 samples, a validation set of 10,000 samples, and a test set of 50,000 samples. We present statistics comparing the original LAION-Aesthetics text-to-image dataset with our LAION-SG dataset as follows. 

As shown in~\cref{tab: statistic feature of dataset}, in the original LAION-Aesthetics caption, the average number of objects per sample is 5.33, with 38\% of these being proper nouns that offer limited guidance during model training. For our SG annotations, the average number of objects per sample increases to 6.39, excluding abstract proper nouns and focusing on specific nouns that reflect true semantic relationships. LAION-SG contains 20\% more object information than the original LAION-Aesthetics dataset, and this advantage increases to 216\% when excluding proper nouns.

Additionally, we calculated the relationship between length and accuracy for different annotations. The annotation length for text is defined as the number of tokens in the prompt, while for SG as the total number of nodes and edges. We leverage SG-IoU+, Entity-IoU+, and Relation-IoU+ introduced in Sec.~\ref{subsec: exp_datasets_metrics} to measure annotation accuracy.

The average annotation length for original captions and our scene graphs is 19.0 and 32.2, respectively, with SG achieving higher accuracy across all three metrics. \cref{fig: scatter} visualizes the length and accuracy of samples for both annotation types. Note that a scene graph is a more structured and compact form of annotation compared to text. Even so, the SG length is still significantly longer than sparse text, and its accuracy is also much higher. This demonstrates that our LAION-SG dataset contains richer, more nuanced, and precise semantic features, which can greatly enhance model performance in image generation, fundamentally addressing the challenges of generating complex scenes. 

Furthermore, we analyze the length distribution of scene graphs in LAION-SG in \cref{fig:bar_donut} (a). Most objects are described by 0-5 (45.72\%) or 5-10 (41.04\%) words, with a smaller proportion described by 10-20 (12.46\%) words or more than 20 (0.79\%) words. This range is reasonable, offering a more precise expression than a single word while avoiding excessive length that could hinder model learning efficiency. 
In terms of the overall scene graph, the proportions of word counts in the ranges 0-10, 10-20, 20-30, and more than 30 are 10.39\%, 32.15\%, 28.80\%, and 28.66\%, respectively. These statistics reflect the richness, detail, and flexibility of annotations in LAION-SG.

\cref{fig:bar_donut} (b) presents the top 10 most frequent relations and attributes in LAION-SG. The most commonly included relation terms are generally specific in semantics like ``surrounded by'', ``adjacent to'' and ``holding''.
The most frequent relation is ``surrounded by'', occurring 80,058 times and accounting for 3.78\% of all relations. The most common attribute is ``tall'', representing 7.36\% of all attributes, while the second most common, ``small'', accounts for only 4.58\%. The tenth-ranked relation and attribute each make up only 1.51\% and 2.2\% of their respective totals. These data indicate the annotations in LAION-SG are highly diverse and broadly covered, as even the most frequently used descriptors represent only a small percentage of the total.

\subsection{Complex Scene Generation Benchmark}
\label{subsec: CompSGen Bench}
To evaluate model performance on compositional image generation, we propose Complex Scene Generation Benchmark (CompSGen Bench). From the 50,000-image test set, we select samples with over four relations as complex scenes, and get a total of 20,838 samples. We calculate FID~\cite{NEURIPS2023_dd83eada_FID} , CLIP score~\cite{Radford2021Learning_CLIP}, and three accuracy metrics~\cite{Shen2024SGAdapterET} to assess models' performance.

FID measures the overall quality of generated images, while the CLIP Score calculates the similarity between the generated and ground truth images. The complex scene evaluation consists of three metrics: SG-IoU, Entity-IoU, and Relation-IoU. They represent the overlap between the generated images and the real annotations in terms of scene graphs, objects, and relations, respectively. Sec.~\ref{subsec: quantitative_results} shows the test results for different models on CompSGen Bench.

\section{Foundation Model}

As the complexity of the image increases, the generated results of T2I models become more difficult to control (\cref{fig: teaser}). We introduce a foundation model to address the challenges of compositional image generation in T2I task. 
Our model is built on top of Stable Diffusion XL (SDXL)~\cite{podell2023sdxlimprovinglatentdiffusion} and incorporates SG information via graph neural networks (GNN)~\cite{scarselli2008graph_GNN}.


An SG consists of multiple triples and single objects. Our baseline initializes each triple and single object separately using the CLIP text encoder \( E_T(\cdot) \). For single objects, the initialization result from CLIP serves as the final representation, denoted as \( \mathbf{e}_s \). For SG triples, each of them is encoded by CLIP to yield a corresponding triple embedding \( \mathbf{e}_t = {E_T}(triple^{sg}) \). Our SG encoder extracts object and relation embeddings as the nodes and edges and inputs them into the GNN to optimize the SG embedding. More calculation details can be found in supplementary material.

If a relation contains multiple words, each word contributes an edge connecting the nodes of the two related objects. Attributes are treated as separate nodes connected to their respective objects. After processing with the GNN, we obtain a refined triple embedding, denoted as \( \mathbf{e}_r \).

To stabilize the training, we introduce a learnable scaling factor \( \alpha \) to control the strength of the refined embedding. \( \alpha \) is initialized as zero and updated throughout training.
Finally, all triple embeddings are concatenated with single-object embeddings to form the SG embedding \( \mathbf e_{sg}\), which is fed into the U-Net of SDXL for iterative noise prediction.
\begin{equation}
    \mathbf e_{sg}=f(sg)=\text {concat}(\mathbf e_t+\alpha \mathbf e_r,\mathbf e_s)
\end{equation}

Given an image \( x \) and its corresponding SG, the SG encoder is trained with:
\begin{equation}
    \mathcal L=\mathbb E_{\mathcal E(x),sg,\epsilon,t}[\parallel \epsilon-\epsilon_\theta(z_t,t,f(sg)) \parallel^2_2]
\end{equation}
Here, \( \mathcal{E}(\cdot) \) denotes the CLIP image encoder, and \( z_t \) represents the latent code of the image. 
\( \epsilon_\theta \) represents the predicted noise, parameterized by \( \theta \). 
\( f(sg) \) encapsulates the SG embedding output from SG encoder of our baseline. 
We train the parameters of SG encoder to minimize the gap between the predicted and added noise, 
thereby fitting the distribution of compositional images.


\begin{figure*}[t]
    \centering
    \includegraphics[width=1\textwidth]{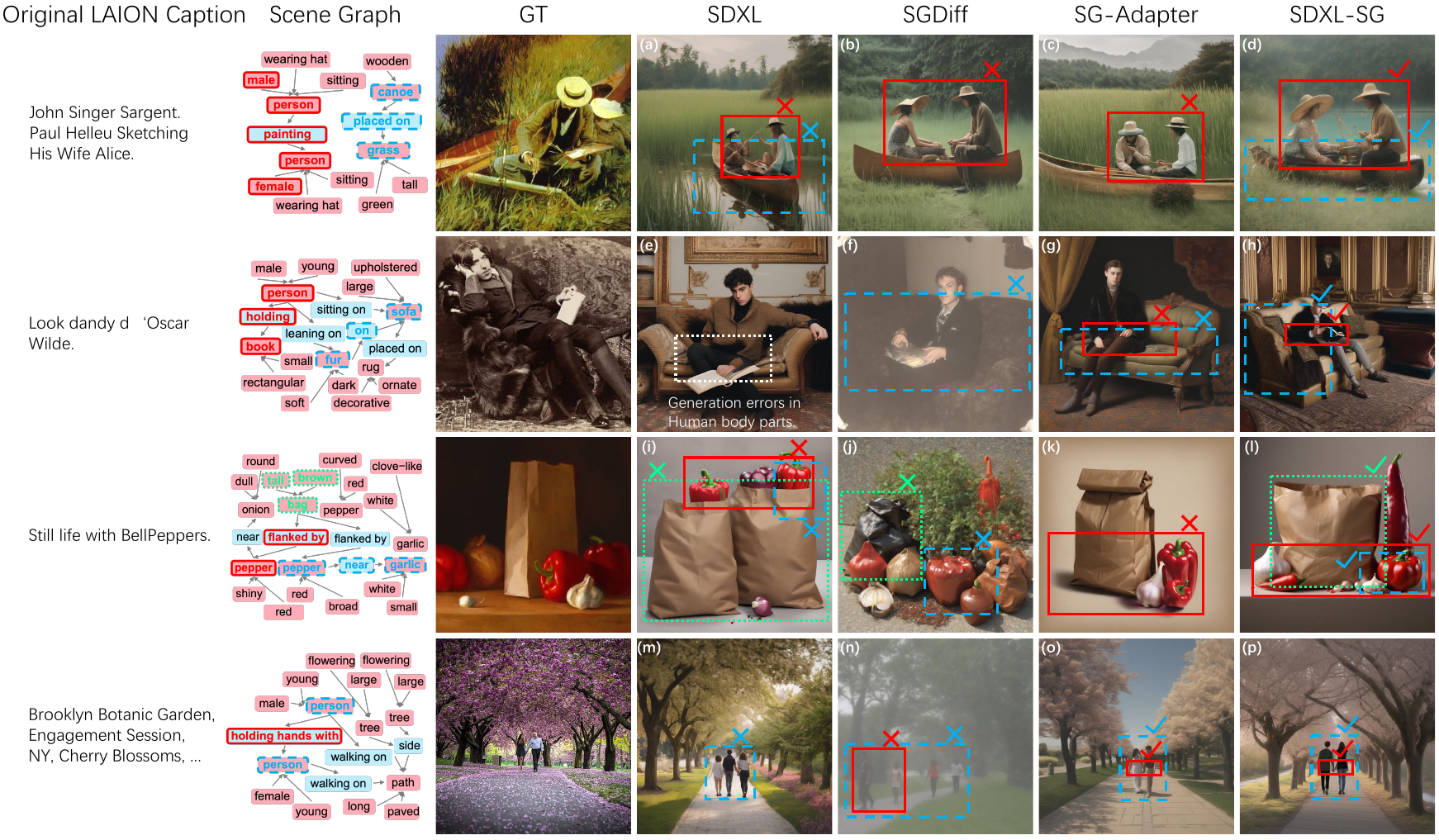} 
    \vspace{-1.5em}
    \caption{Visual comparison on LAION-SG. The compared methods include T2I model (SDXL~\cite{podell2023sdxlimprovinglatentdiffusion}) and SG2IM models (SGDiff~\cite{Yang2022DiffusionBasedSG} and SG-Adapter~\cite{Shen2024SGAdapterET}). The first column shows the original caption from LAION-Aesthetics. The second column displays the scene graph from our LAION-SG. The last five columns show ground truth images and images generated by different models. Objects or relations are highlighted with the same color in scene graphs and generated images to show SDXL-SG successfully captures the complex scenes.}
    \label{fig: qualitative}
    \vspace{-1em}
\end{figure*}

\section{Experiments}

\subsection{Implementation and Baselines}
We compare our trained SDXL-SG with three advanced baselines: SDXL~\cite{podell2023sdxlimprovinglatentdiffusion}, SG-Adapter~\cite{Shen2024SGAdapterET}, and SGdiff~\cite{Yang2022DiffusionBasedSG}, following their evaluation settings. For SDXL-SG, we initialize the scene graph embeddings using OpenCLIP ViT-bigG/14~\cite{ilharco2021OpenCLIP} and CLIP ViT-L/14~\cite{Radford2021Learning_CLIP}, similar to SDXL. These embeddings are refined with a 5-layer SG Encoder, each with 512 input and output dimensions. The refined embeddings are subsequently injected into SDXL for image generation. For our training, we employ the Adam optimizer with a learning rate of 5e-4, training for one epoch on the full LAION-SG dataset. All training processes are conducted on eight NVIDIA RTX 4090D GPUs.

\subsection{Datasets and Evaluation Metrics}
\label{subsec: exp_datasets_metrics}
We train existing models and SDXL-SG on COCO-Stuff, Visual Genome (VG), and LAION-SG datasets. 
We evaluate the generated results based on image quality and accuracy in target content within complex scenes. Fréchet Inception Distance (FID)~\cite{NEURIPS2023_dd83eada_FID} measures the overall visual quality of generated images, while the CLIP Score evaluates the similarity between generated and ground truth (GT) images. For assessing complex scenes, we calculate SG-IoU, Entity-IoU, and Relation-IoU~\cite{Shen2024SGAdapterET}.
They respectively indicate the degree of consistency between the scene graphs, objects and relations derived from the generated image and the real annotation.
Higher scores indicate stronger model capability in generating complex scenes.

To evaluate the annotation quality, we propose SG-IoU+, Entity-IoU+, and Relation-IoU+. 
Images are generated using scene graphs or LAION captions. The SG list, entity list, and relation list are then extracted from both the generated and GT images. The consistency between the corresponding lists of the two images is calculated to assess the annotation accuracy. The extraction is performed using GPT-4o. Given the high cost, we compute the average for 300 samples as the result. More calculation details can be found in the supplementary material.


\subsection{Qualitative Results}
\cref{fig: qualitative} displays 1024\(\times\)1024 images generated on LAION-SG. Each row shows an original caption from the LAION-Aesthetics dataset, a scene graph, the corresponding ground truth image, and images generated by different baseline models. The corresponding elements in the scene graph and the images are highlighted in matching colors. 

Our foundation model SDXL-SG can generate scenes with more accurate objects and relations, even for relatively complex scenarios. For instance, in the first row, where the relationship is ``male person painting female person'', both (a) and (b) fail to generate ``painting'', and (c) generates two female figures, whereas SDXL-SG accurately and qualitatively generates the relations provided by the scene graph. Figures (f)-(p) illustrate more examples where ours outperforms existing baselines in relation generation. 

Additionally, existing T2I and SG2IM models more frequently produce incorrect generations in (e). Other errors include erroneous number of generated objects such as bag in the green box in (i), person in the blue box in (m) and (n) or attribute errors such as bag in the yellow box in (j). SDXL-SG avoids these mistakes as in (d), (h), (l) and (p).

\begin{table}
    \centering
    \begin{tabular}{@{}lccccc@{}}
    \toprule
        \textbf{Method} & \textbf{Dataset} & \textbf{FID}$^\downarrow$ & \makecell{\textbf{SG-}\\ \textbf{\textbf{IoU}}$^\uparrow$} & \makecell{\textbf{Ent.-}\\ \textbf{IoU}$^\uparrow$} & \makecell{\textbf{Rel.-}\\ \textbf{IoU}$^\uparrow$} \\ 
        \midrule
        SDXL & LAION & \textbf{19.3} & 0.371 & 0.813 & 0.780 \\
        \midrule
        \multirow{3}{*}{\makecell[l]{SGDiff\\w/o bbox}} & COCO & 47.8 & 0.435 & 0.841 & 0.816 \\
        & VG & 35.2 & 0.529 & 0.801 & 0.795 \\
        & \makecell{LS} & 32.2 & 0.531 & 0.855 & 0.830 \\ 
        \midrule
        \multirow{3}{*}{\makecell[l]{SG-\\Adapter}} & COCO & 34.9 & 0.485 & 0.840 & 0.833 \\ 
        & VG & 39.5 & 0.515 & 0.803 & 0.782 \\ 
        & \makecell{LS} & 31.3 & 0.538 & \underline{0.866} & \underline{0.852} \\ 
        \midrule
        \multirow{3}{*}{\makecell[l]{SDXL-\\SG (Ours)}} & COCO & 30.0 & 0.497 & 0.842 & 0.833 \\
        & VG & 21.9 & \underline{0.546} & 0.813 & 0.800 \\ 
        & \makecell{LS} & \underline{20.1} & \textbf{0.558} & \textbf{0.884} & \textbf{0.856} \\
        \bottomrule
    \end{tabular}
    \vspace{-0.5em}
    \caption{Results on COCO-Stuff, Visual Genome and LAION-SG (LS). The first and second best is in \textbf{bold} and \underline{underlined}.}
    \label{tab: quantitative}
    \vspace{-1em}
\end{table}

\begin{table}
    \centering
    \begin{tabular}{@{}lccccc@{}}
    \toprule
        \textbf{Method} & \textbf{FID}$^\downarrow$ & \makecell{\textbf{CLIP}$^\uparrow$} & \makecell{\textbf{SG-}\\ \textbf{IoU$^\uparrow$}} & \makecell{\textbf{Ent.-}\\ \textbf{IoU}$^\uparrow$} & \makecell{\textbf{Rel.-}\\ \textbf{IoU}$^\uparrow$} \\
        \midrule
        SDXL & \textbf{25.2} & \textbf{0.700} & 0.226 & 0.753 & 0.658 \\ 
        SGDiff & 35.8 & 0.690 & 0.304 & \underline{0.787} & \underline{0.698} \\ 
        \makecell[l]{SG-Adapter} & 27.8 & 0.681 & \underline{0.314} & 0.771 & 0.693 \\ 
        \makecell[l]{SDXL-SG\\(Ours)} & \underline{26.7} & \underline{0.698} & \textbf{0.340} & \textbf{0.792} & \textbf{0.703} \\ 
        \bottomrule
    \end{tabular}
    \vspace{-0.5em}
    \caption{The results of existing T2I and SG2IM models, as well as our baseline model, on the Complex Scene Generation Benchmark. The best is in \textbf{bold}, and the second best is \underline{underlined}.}
    \label{tab: complex benchmark}
    \vspace{-1em}
\end{table}

\subsection{Quantitative Results}
\label{subsec: quantitative_results}
We compared results of SDXL~\cite{podell2023sdxlimprovinglatentdiffusion}, other SG2IM methods including SGDiff~\cite{Yang2022DiffusionBasedSG} and SG-Adapter~\cite{Shen2024SGAdapterET}, and SDXL-SG. Each is tested when trained on different datasets. The original SGDiff~\cite{Yang2022DiffusionBasedSG} introduces bounding box as auxiliary data during training. For fair comparison, we train SGDiff without bounding box with the official implementation. We used FID
to evaluate the quality of generated images. Fine-tuning pre-trained T2I models inevitably increases FID scores~\cite{DreamBooth,Shen2024SGAdapterET,wang2024distributedevaluationgenerativemodels}. We also measure SG-IoU, Entity-IoU, and Relation-IoU~\cite{Shen2024SGAdapterET}.

As demonstrated in \cref{tab: quantitative}, our baseline achieves the best performance among all candidates in both image quality and accuracy. Notably, the SG-IoU of T2I model is significantly lower than that of SG2IM models, indicating that text provides far less control in the image generation process compared to structured annotations. This highlights the necessity of constructing a large-scale, high-quality structured annotation dataset. Furthermore, for the same model, results trained on LAION-SG consistently outperformed those trained on COCO and VG. This suggests that our LAION-SG dataset is more effective than previous SG-image datasets due to its higher annotation quality.

Additionally, we evaluate the complex scene generation capability of baseline models on the CompSGen Bench (Sec.~\ref{subsec: CompSGen Bench}). As shown in~\cref{tab: complex benchmark}, our baseline outperforms the existing SG2IM model in terms of image quality, similarity to the ground truth (GT) image, and content accuracy. Compared to SDXL, our FID score does not increase significantly even after fine-tuning, which typically raises FID scores. The slightly lower CLIP score (only by 0.02) compared to SDXL is due to pre-trained CLIP models incorporate abstract prior information such as background and historical knowledge. In contrast, our focus is on generating accurate and realistic content for complex scenes. SDXL-SG significantly outperforms SDXL on accuracy metrics
including SG-IoU, Entity-IoU and Relation-IoU.

We also conduct a quantitative analysis (detailed in Sec.~\ref{subsec: LAION-SG dataset} ) and a user study (detailed in supplementary material) to verify the effectiveness and strong correlation with human perception of structured annotations, indicating their significant advantage in expressing image content compared to sequential text.


\subsection{Ablation Study}
We conduct ablation studies to demonstrate the positive impact of LAION-SG. We train SDXL-SG variants on 10\%, 20\%, 50\%, and 100\% samples of LAION-SG. 
The total training iterations remain constant across all settings for fairness. As the sample size increases, the model’s capability to generate compositional images improves significantly (\cref{tab: ablation}). Notably, in the 10\% LAION-SG ablation, where the data volume is smaller than that of VG, the model’s FID and Entity-IoU scores still outperform the results trained on VG, with SG-IoU and Relation-IoU scores remaining roughly comparable (\cref{tab: quantitative}). This indicates LAION-SG not only provides a data volume advantage but also features higher quality in images and annotations.
This enhances model training efficiency and significantly improves performance in compositional image generation.
\begin{table}
  \centering
  \begin{tabular}{@{}lccccc@{}}
    \toprule
    \textbf{Method} & \makecell{\textbf{Prop.}} & \textbf{FID}$^\downarrow$ & \makecell{\textbf{SG-}\\ \textbf{IoU}$^\uparrow$} & \makecell{\textbf{Ent.-}\\ \textbf{IoU}$^\uparrow$} & \makecell{\textbf{Rel.-}\\ \textbf{IoU}$^\uparrow$} \\
    \midrule
    \multirow{4}{*}{\makecell[l]{SG-\\Adapter}} & 10\% & 31.6 & 0.522 & 0.794 & 0.790 \\
    & 20\% & 24.3 & 0.524 & 0.804 & 0.793 \\
    & 50\% & 22.9 & 0.535 & 0.800 & 0.796 \\
    & 100\% & 21.9 & 0.546 & 0.813 & 0.800 \\
    \midrule
    \multirow{4}{*}{\makecell[l]{SDXL-\\SG (Ours)}} & 10\% & 27.3 & 0.530 & 0.874 & 0.837 \\
    & 20\% & 24.5 & 0.533 & 0.877 & 0.838 \\
    & 50\% & 22.2 & 0.547 & 0.876 & 0.849 \\
    & 100\% & \textbf{20.1} & \textbf{0.558} & \textbf{0.884} & \textbf{0.856} \\
    \bottomrule
  \end{tabular}
  \vspace{-0.5em}
  \caption{Results of ablation study. Prop. denotes data proportion.}
  \label{tab: ablation}
  \vspace{-1em}
\end{table}

\section{Conclusion}
In this paper, we introduce LAION-SG, a high-quality, large-scale dataset with structured annotation designed for training complex image-text models. Compared to existing text-image datasets, LAION-SG provides more precise descriptions of objects, attributes, and relations, effectively capturing the semantic structure of complex scenes. 
Based on LAION-SG, we train a new foundational model SDXL-SG, integrating structured annotation into the generation process. Experiments demonstrate models trained on LAION-SG significantly outperform those trained on existing datasets.
We also propose CompSG-Bench, a benchmark for evaluating model performance in compositional image generation based on scene graph, establishing a new standard for complex scene generation. 
In summary, LAION-SG represents a pioneering effort in annotating complexity on image datasets and holds great potential for broader scene perception and synthesis applications.

%% file: sec/mcy_SM.tex
\renewcommand{\thesection}{S\arabic{section}}
\renewcommand{\thetable}{S\arabic{table}}
\renewcommand{\thefigure}{S\arabic{figure}}
\renewcommand{\theequation}{S\arabic{equation}}

\begin{figure*}[t]
    \centering
    \includegraphics[width=1\textwidth]{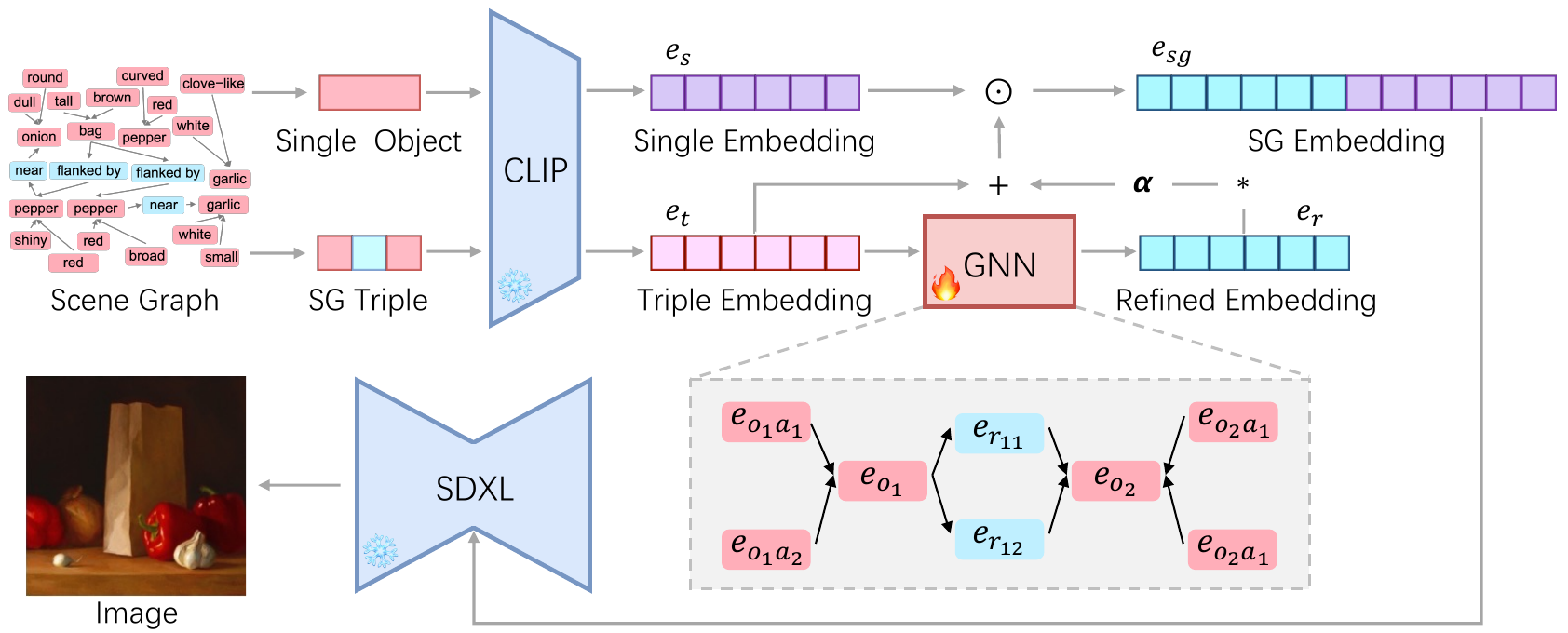} 

    \caption{The architecture of our foundation model. Concatenation is indicated by $\odot$ and multiplication by $\ast$.}
    \label{fig: model}
\end{figure*}

\setcounter{section}{0}
\section{Details of SDXL-SG}

Text-to-image (T2I) generation can produce highly detailed results. However, when the given text describes a relatively complex scene (e.g., an image containing multiple objects or multiple relationships between objects), the output of T2I models often falls short of expectations. Through experiments, we found that incorporating scene graph (SG) information during the image generation can significantly improve the model's ability to generate compositional images. Based on this observation, we introduce a foundation model to alleviate restrictions on text-to-image generation. 

Our model is based on the SDXL~\cite{podell2023sdxlimprovinglatentdiffusion} architecture, integrating SG information into the generation process through graph neural networks (GNN)~\cite{scarselli2008graph_GNN}. As shown in ~\cref{fig: model}, we initialize each single object and triple of SG separately using the CLIP text encoder \( E_T(\cdot) \) to get their embeddings $e_s$ and $e_t$. Specifically, for the triple embedding, it includes representations of objects \( \mathbf{e}_{o_k} \), relations \( \mathbf{e}_{r_{ij}} \), and object attributes \( \mathbf{e}_{o_n a_m} \). Here, \( \mathbf{e}_{o_k} \) represents the embedding of the \( k \)-th object in the SG, \( \mathbf{e}_{r_{ij}} \) represents the embedding of the \( j \)-th word in the \( i \)-th relation, as some relations in LAION-SG annotations may contain multiple words (e.g., ``grown by''), and \( \mathbf{e}_{o_n a_m} \) denotes the embedding of the \( m \)-th attribute word of the \( n \)-th object, as an object’s attributes may consist of multiple words (e.g., ``tall wooden building'').

This structured SG input is then fed into the GNN. Objects serve as nodes, and relations act as edges. After that, we obtain a refined triple embedding \( \mathbf{e}_r \), which can be represented as:
\begin{equation}
    \mathbf e_r=\text {GNN}({E_T}(triple^{sg}))
\end{equation}
We introduce an $\alpha$ factor to control the strength of the refined triple embedding, ensuring stable learning for the model. The optimized triple embedding is represented as:
\begin{equation}
    \mathbf e_{t^\prime}=\mathbf e_t+\alpha \mathbf e_r
\end{equation}
Finally, all triple embeddings are concatenated with single-object embeddings to form the SG embedding \( \mathbf e_{sg}\), which is fed into the U-Net of SDXL for iterative noise prediction.
\begin{equation}
    \mathbf e_{sg}=f(sg)=\text {concat}(\mathbf e_{t^\prime},\mathbf e_s)
\end{equation}
We employ SDXL~\cite{podell2023sdxlimprovinglatentdiffusion} as the pretrained framework. The model learns SG knowledge at time step \( t \) by:
\begin{equation}
    \mathcal L=\mathbb E_{\mathcal E(x),sg,\epsilon,t}[\parallel \epsilon-\epsilon_\theta(z_t,t,f(sg)) \parallel^2_2]
\end{equation}
As introduction in ~\cref{subsec: SDXL}, our training is conducted in the latent space to enhance efficiency. \( f(sg) \) encapsulates the SG embedding output from SG encoder of our baseline. 
The training process dynamically adjusts parameters of SG encoder to minimize the gap between the predicted and added noise, which can reduce \( \mathcal{L} \), improving the model’s capability to handle compositional image generation.

Our architecture is designed to be lightweight and efficient. The generation time for 100 images at a resolution of 1024$\times$1024 is measured. Our baseline model takes an average of 17.19 seconds per image, while the original SDXL model takes 16.70 seconds, both running on a single RTX 4090D GPU. 
Moreover, our SgEncoder model has a parameter count of 14.70M, which is only 0.23\% of the approximately 6.6B parameters of the original SDXL, demonstrating its exceptional lightweight advantage.
The inference time increases by less than 3\%, and the parameter growth is negligible, making the additional computational cost almost insignificant. However, the improvement in output accuracy is substantial.

\section{Discussion on Complex Scene Generation}

Complex scene generation is a challenging task attracting attention from the community of image generation. MIGC (Multi-Instance Generation Controller)~\cite{zhou2024migc} and MIGC++~\cite{zhou2024migc++} adopts a strategy of generating individual instances separately and then integrating them, while incorporating multimodal descriptions for attributes (text and images) and localization (bounding boxes and masks).
By incorporating appearance tokens and an instance semantic map, IFAdapter~\cite{wu2024ifadapter} enhances the fidelity of fine-grained features in multi-instance generation while ensuring spatial precision. 
3DIS~\cite{zhou20243dis} decouples the multi-instance generation task into two stages: depth map generation and detail rendering. By combining depth-driven layout control with training-free fine-grained attribute rendering, it significantly enhances instance positioning accuracy and detail representation. 

These works effectively address the challenges of multi-instance compositional generation. However, they primarily control the generated objects at the spatial level and fail to resolve inaccuracies in generating abstract semantic relationships between objects, such as ``holding'' or ``riding''.

In contrast, \cite{wang2024scene} disentangles layouts and semantics from scene graphs, leveraging variational autoencoders and diffusion models to significantly enhance instance relationship modeling and fine-grained control in complex scene generation. This approach enhances the model's understanding and representation of abstract semantics through the structured form of scene graphs. Nevertheless, it still faces generation bottlenecks due to dataset limitations. 

Therefore, we propose the LAION-SG dataset to fundamentally address the challenges of complex scene generation at the data level. Simultaneously, we introduce a baseline model that enables simple and efficient generation of complex scenes based on scene graphs.

\section{User Study}
\begin{figure}[t]
    \centering
    \includegraphics[width=0.47\textwidth]{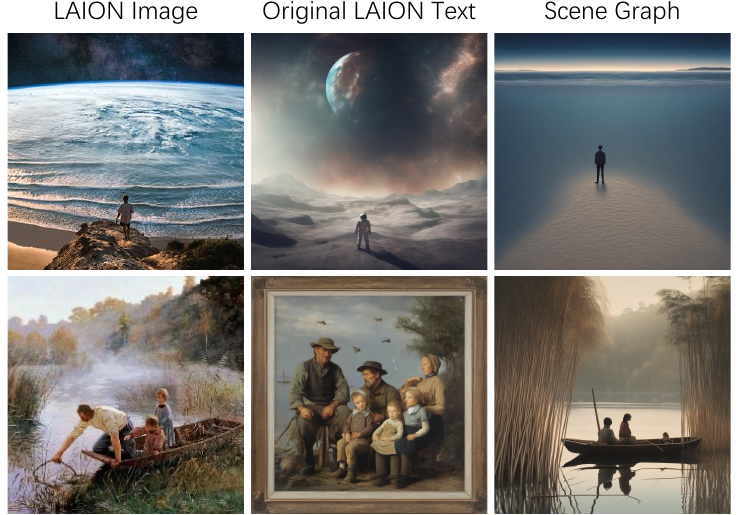} 
    
    \vspace{-0.5em} 
    
    \caption*{\textit{Please select the image closest to the LAION image from those generated from the original LAION caption and the scene sraph.}}
    \begin{tabular}{@{}lcc@{}}
        \toprule
        Annotation & Original Caption & Scene Graph \\
        \midrule
        User Preference & 37\% & 63\% \\
        \bottomrule
    \end{tabular}
    \caption{The result of user study. \textbf{Top:} We present images generated from original captions and scene graphs to users and ask them to choose the one that better aligns with the content of the LAION image. \textbf{Bottom:} Across 100 validation image pairs, users showed a strong preference for the results generated from scene graphs.}
    \label{fig: user_study}
\end{figure}
Beyond objective metrics, whether the results align with human cognition is also crucial. We conduct a user study to compare which annotation type generates images that better align with human perception.

We randomly select 100 text-sg-image triplets. In each trial, users are presented with three images: the LAION image and two images generated from the original LAION caption and the scene graph respectively. Users are asked to choose the image from the latter two that best matched the content of the LAION image. We invite 10 participants, with a 1:1 gender ratio and ages ranging from 20 to 30. They come from diverse backgrounds, including computer science, design, and human-computer interaction (HCI).

The result of user study is shown in ~\cref{fig: user_study}. A total of 63\% of participants preferred the images generated from the scene graph, while only 37\% chose those from the text prompt. 
This indicates that, compared to sequential text annotations, structured annotations have an overwhelming advantage in expressing image content.

\textbf{Notification to Human Subjects.} We present the notification to subjects to inform the collection and user of data before the experiments.

\begin{quote}
    
Dear volunteers, we would like to express our thankfulness for your support to our study. We study an image generation algorithm, which translates scene graphs to realistic images.

All information about your participation in the study will appear in the study record. All information will be processed and stored according to the local law and policy on privacy. Your name will not appear in the final report. When referred to your data provided, only an individual number assigned to you is mentioned.

We respect your decision whether you want to be a volunteer for the study. If you decide to participate in the study, you can sign this informed consent form.
\end{quote}

The use of users' data was approved by the Institutional Review Board of the main authors' affiliation.

\section{Examples of LAION-SG}
\begin{figure}[t]
    \centering
    \includegraphics[width=0.47\textwidth]{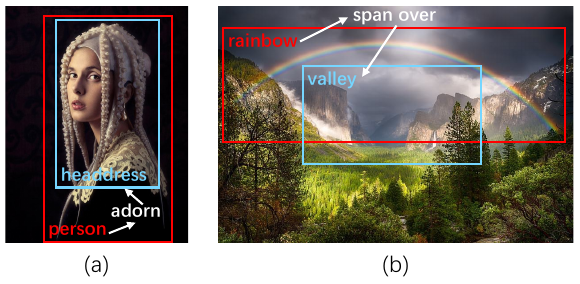} 

    \caption{Example images from the LAION-Aesthetics dataset.}
    \label{fig: example_imgs}
\end{figure}

Given an image, we employ a multimodal large language model, GPT-4o~\cite{openai2024gpt4technicalreport}, to perform automated scene graph annotation. Our pipeline focuses on assigning distinct ids to different objects, identifying attributes for each object, labeling relations between objects, and adhering to other specified constraints. The annotations are strictly output in the designated format. Here, we provide two specific examples. For the image in ~\cref{fig: example_imgs} (a), the highlighted portion corresponds to the following scene graph, with other parts omitted.

\begin{lstlisting}
{
    "img_id": "482063",
    "name": "minus83166520...",
    "caption_ori": "Page 90 of Girl...",
    "score": "6.720815181732178",
    "url": "https://stories...",
    "items": [
        {
            "item_id": 0,
            "label": "person",
            "attributes": [
                "young",
                "female"
            ],
            "global_item_id": 3201686
        },
        {
            "item_id": 1,
            "label": "headdress",
            "attributes": [
                "ornate",
                "white"
            ],
            "global_item_id": 3201687
        },
        ...
    ],
    "relations": [
        {
            "triple_id": 0,
            "item1": 1,
            "relation": "adorn",
            "item2": 0,
            "global_relation_id": 2118510
        },
        ...
    ]
},
\end{lstlisting}
And for the image in ~\cref{fig: example_imgs} (b), its highlighted portion corresponds to the following scene graph, with other parts omitted.

\begin{lstlisting}
{
    "img_id": "483868",
    "name": "694108219422834467.jpg",
    "caption_ori": "Yosemite's Rainbow.  Yosemite National Park, California.",
    "score": "6.544332504272461",
    "url": "https://photos.smugmug.com/..."
    "items": [
        {
            "item_id": 0,
            "label": "rainbow",
            "attributes": [
                "colorful",
                "arc-shaped"
            ],
            "global_item_id": 3213781
        },
        ...
        {
            "item_id": 4,
            "label": "valley",
            "attributes": [
                "green",
                "vast"
            ],
            "global_item_id": 3213785
        },
        ...
    ],
    "relations": [
        {
            "triple_id": 0,
            "item1": 0,
            "relation": "span over",
            "item2": 4,
            "global_relation_id": 2126675
        },
        ...
    ]
},
\end{lstlisting}

\section{Details of Accuracy Metrics}

We leverage SG-IoU, Entity-IoU, and Relation-IoU~\cite{Shen2024SGAdapterET} to measure the model's ability to generate complex scenes. Specifically, we use GPT-4 to extract scene graph lists from the generated images, with each list consisting of triples in the form \( \langle s_n, r_n, o_n \rangle \). From this SG list, we derive the Entity and Relation lists and calculate the intersection over union (IoU) between the derived lists and the real annotations. 
Higher scores indicate stronger model capability in generating complex scenes.

Furthermore, we propose SG-IoU+, Entity-IoU+, and Relation-IoU+ to evaluate the annotation accuracy. 
Detailedly, we first generate two images: one using the original LAION captions and the other using scene graph from LAION-SG.
Then for the real image and the two generated images, we extract the lists of SGs, relations and entities from each image with GPT-4o again. 
Taking the lists of SGs as an example, the IoU scores is calculated between the list SG generated image and that of the real image. Also the IoU beween the caption generated and the real is calculated.
This IoU evaluates the extent that the generated images and the real image are similar along the SG structure, thus reflecting the annotation accuracy.
It is the hight the better.
Such IoU is also calculated on the lists of relations and entities.


\section{Discussion on Annotation}
\subsection{Hallucinations of GPT-4o}
\label{subsec: Hallucinations of GPT-4o}
\begin{figure}[t]
    \centering
    \includegraphics[width=0.47\textwidth]{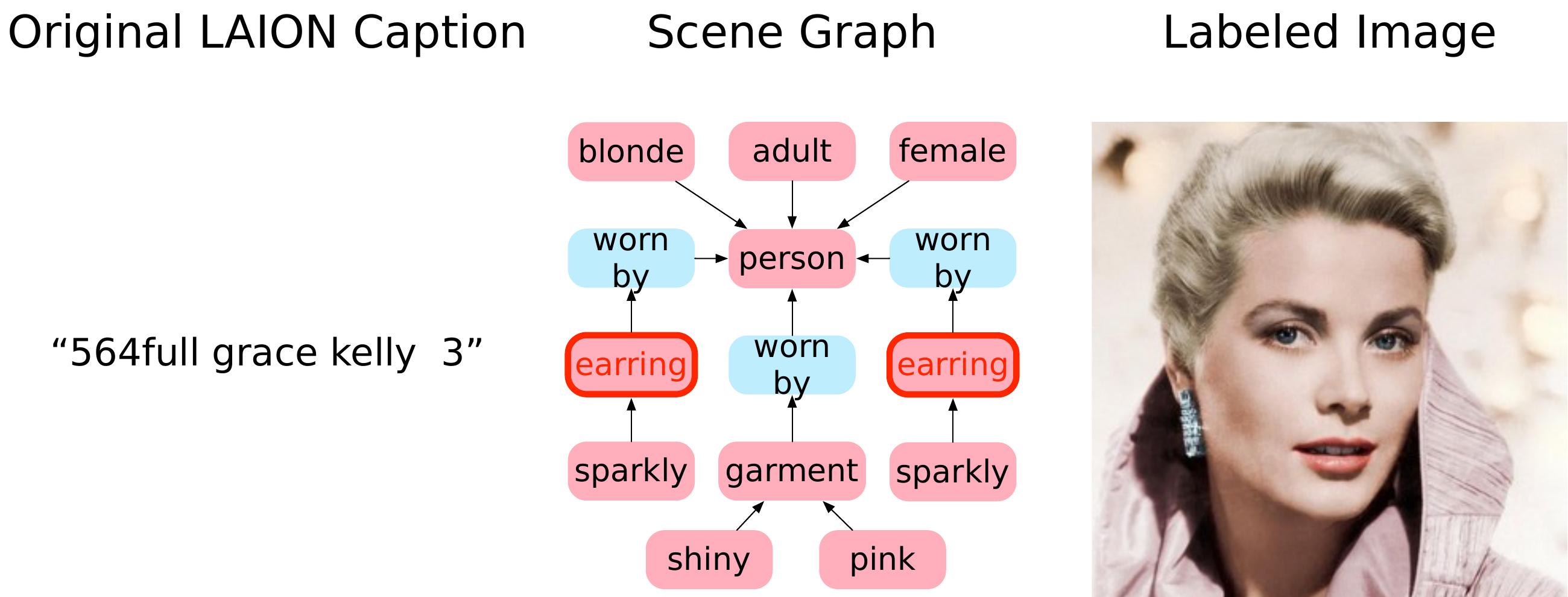} 

    \caption{GPT-4o occasionally exhibits hallucination phenomena, labeling objects that do not exist in the image. For example, in a case where the image shows only one earring, GPT-4o incorrectly labels a nonexistent second earring. However, despite these issues, the overall quality of our annotations still surpasses that of LAION's original annotations.}
    \label{fig: earrings_img}
\end{figure}

In our annotation process, GPT-4o occasionally exhibits hallucination phenomena, generating information that does not actually exist. Through a random check of 100 annotation samples, we find that approximately 1\% contain such issues. These issues typically manifest as annotations that do not strictly adhere to the image content but instead rely on semantic inference to incorrectly label objects that are not present. For example, in~\cref{fig: earrings_img}, the GT image only shows one visible earring, while the other earring is occluded. However, GPT-4o erroneously infer its presence based on semantic reasoning.

Although the limitations of current multimodal large models make it challenging to completely avoid such problems, the quality of the original LAION annotation of the GT image in Fig. \ref{fig: earrings_img} is relatively low, further hindering the generation of complex scenes. Nevertheless, our annotation process strives to ensure the accurate description of entities and relationships within images, thereby maintaining a high overall annotation quality.

\subsection{Failure Examples}
\label{subsec: Failure Examples}
\begin{figure}[t]
    \centering
    \includegraphics[width=0.47\textwidth]{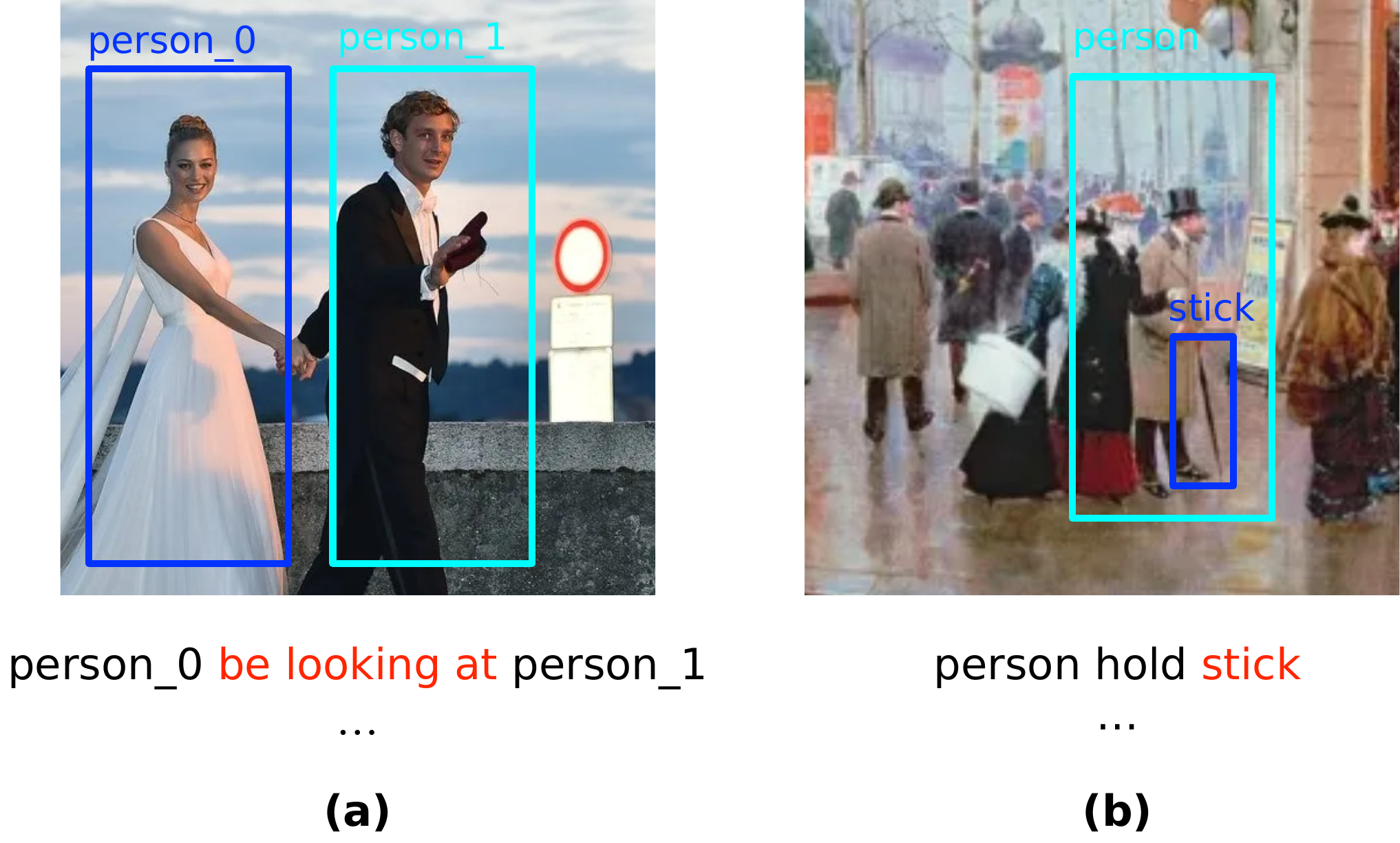} 

    \caption{GPT-4o occasionally makes errors during the annotation process. For example, in (a), GPT-4o misidentifies the relationship, incorrectly assuming that person\_0 is looking at person\_1, which in fact is wrong. In (b), GPT-4o misclassifies a small object, labeling the item held by the person as a stick, when it is actually an umbrella.}
    \label{fig: wrong_img}
\end{figure}

Through prompt engineering, we leverage GPT-4o to perform large-scale, high-quality scene graph annotations for images. While the majority of these annotations accurately describe the entities and their relationships in the images, a small number of errors still occur. In a random sample of 100 annotations, approximately 2\% contain one mislabeled annotation, including inaccurate relationship descriptions (as shown in Fig. \ref{fig: wrong_img}(a)) or entity recognition errors (as shown in Fig. \ref{fig: wrong_img}(b)). 

\subsection{Abstract Attribute}
We follow the guidelines of previous work~\cite{krishna2017visual} and try to ensure that attributes in the scene graph are abstract adjectives rather than specific objects.

\section{More Information of Foundation Model}

\subsection{A Brief Introduction of Stable Diffusion XL}
\label{subsec: SDXL}
SDXL (Stable Diffusion XL)~\cite{podell2023sdxlimprovinglatentdiffusion} is an advanced latent diffusion model (LDMs)~\cite{ldm} primarily designed for generating high-resolution images based on text prompts. It builds on the fundamentals of diffusion models~\cite{ddpm} by utilizing a two-stage process: initially generating images from noise and then refining these images to enhance quality.

SDXL operates in a compressed latent space rather than the pixel space directly, using an autoencoder to encode an input image into a lower-dimensional latent space and then applying the diffusion process in this space. This approach is computationally efficient and enables the generation of high-quality, detailed images with fewer resources compared to pixel-based diffusion models~\cite{ddpm}. The model’s architecture consists of an autoencoder and a UNet-based diffusion network that performs the denoising operations.

To interpret text prompts with high fidelity, SDXL integrates two text encoders (OpenCLIP ViT-bigG~\cite{ilharco2021OpenCLIP} and CLIP ViT-L~\cite{Radford2021Learning_CLIP}). These encoders convert the textual input into feature representations, which are then concatenated and used to condition the diffusion process, thereby allowing the model to follow text prompts more accurately.

The SDXL diffusion process is a series of denoising steps in which the model progressively reduces noise from an initial noise-filled image until a clear image is produced. This iterative process can be represented mathematically by
\begin{equation}
      x_ {t}  =  \sqrt {\alpha _ {t}}   \cdot   x_ {0}  +  \sqrt {1-\alpha _ {t}}   \cdot   \epsilon  
\end{equation}
Here $x_t$ is the noisy image at step $t$. $\alpha_t$ is a noise decay factor for each time step. $x_0$ represents the clean, noise-free image. And $\epsilon$ is random Gaussian noise added to the image at each step. Each step is controlled by a learned model, $\epsilon_\theta$, that predicts and subtracts noise from the image, allowing it to converge on a high-quality result as $t \rightarrow 0$.

SDXL’s training objective is to minimize the mean squared error between the predicted noise and the actual noise added to the image. The conditional term is introduced through classifier-free guidance~\cite{ho2021classifier}, a mechanism that combines conditional information with the noise predictions from unconditional generation. This enables the model to better follow prompt details when generating images.

Specifically, the conditional loss function in SDXL can be represented as
\begin{equation}
       \mathcal L = \mathbb E_{\mathcal E(x_ {0}) , c,  \epsilon   \sim  N(0,I),t}[\parallel \epsilon - \epsilon_{\theta } (z_t, t, \tau (c))\parallel_{2}^{2}],
\end{equation}

where $\mathcal E(x_ {0})$ and $z_t$ is latent representations of the original image and its noisy version at timestep $t$, $c$ and $\tau (c)$ is the input condition and its latent embedding, and $\epsilon_{\theta } (z_t, t, \tau (c))$ represents the model’s noise prediction under condition $c$. Additionally, the conditional term $c$ includes spatial conditions like size and crop settings, enabling the model to adapt to various resolutions and framing needs. By minimizing this error, SDXL learns how to progressively remove noise and refine images accurately across various levels of initial noise.

To enhance the visual quality of generated images, SDXL includes a refinement model that operates in the latent space. This model further refines the output using SDEdit~\cite{meng2022sdedit}, an image-to-image process where noise is temporarily reintroduced and then denoised to improve quality.

\subsection{Additional Results}

\begin{figure*}[t]
    \centering
    \includegraphics[width=1\textwidth]{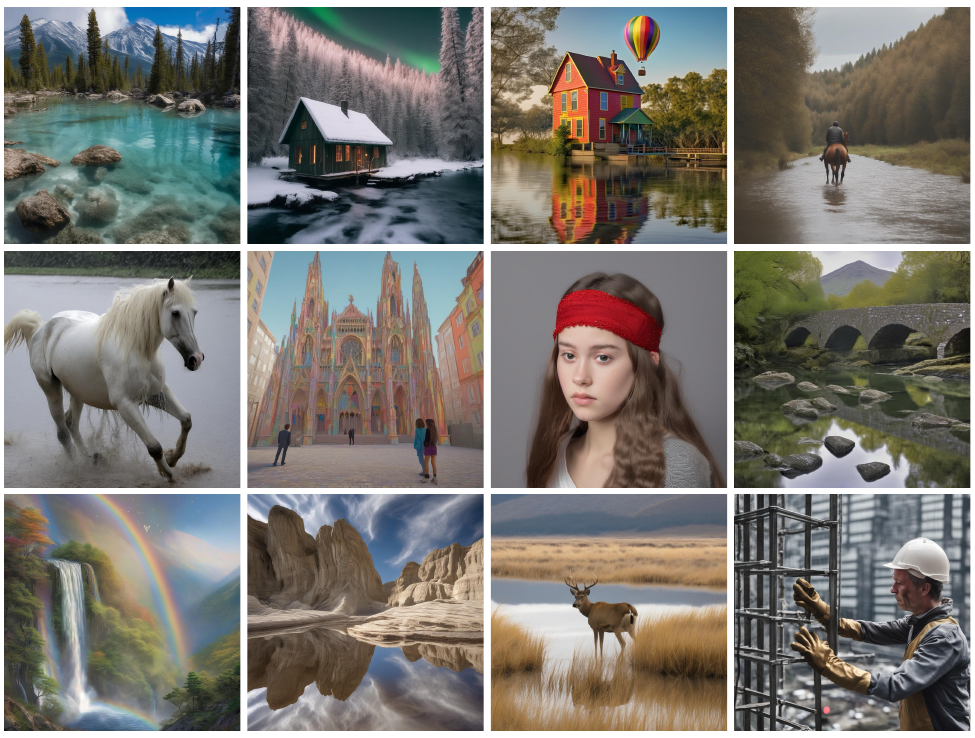} 
    \caption{Additional results of successful examples.}
    \label{fig: results_success}
\end{figure*}

\begin{figure*}[t]
    \centering
    \includegraphics[width=1\textwidth]{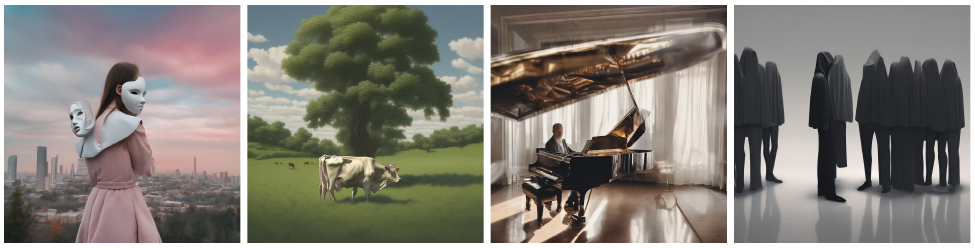} 
    \caption{Additional results of failure examples.}
    \label{fig: results_failure}
\end{figure*}

\begin{figure*}[t]
    \centering
    \includegraphics[width=1\textwidth]{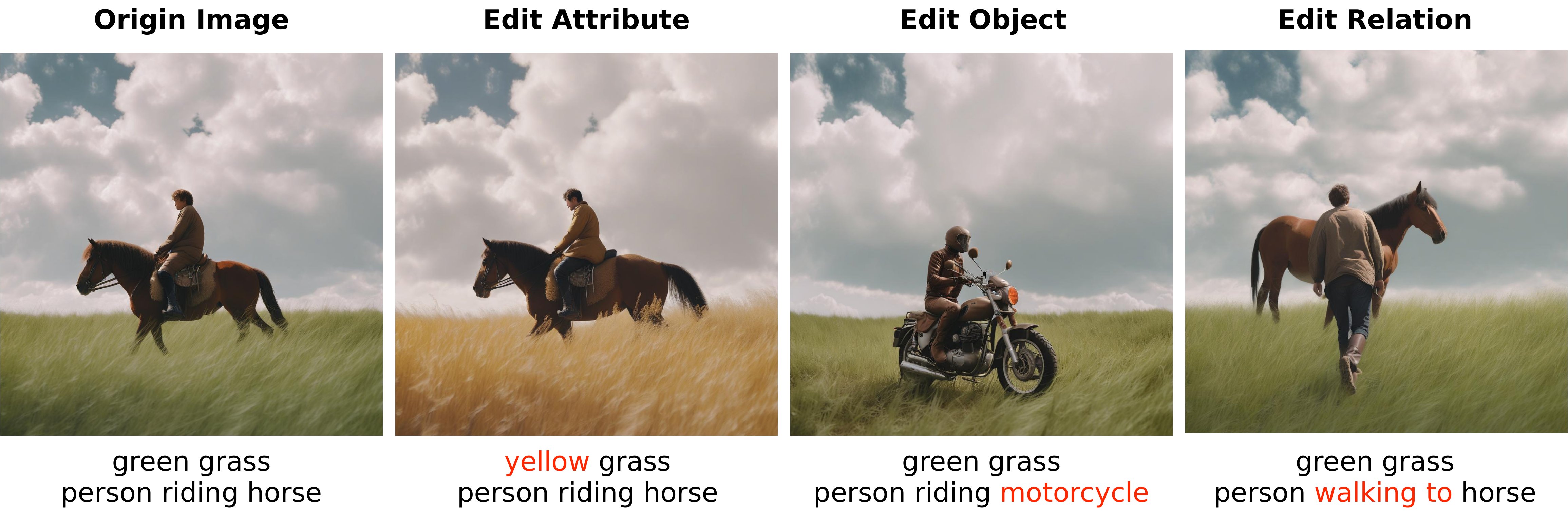} 
    \caption{Additional results of applications in image editing: achieved by editing attributes, objects, or relationships in the scene graph.}
    \label{fig: edit_img}
\end{figure*}

\textbf{Successful and Failure Examples.} We provide additional experimental results. \cref{fig: results_success} presents more successful cases, while \cref{fig: results_failure} shows some failure cases, including object misalignment, incorrect object shape generation, and errors in object appearance generation.

\textbf{Applications in Image Editing.} 
We also offer a straightforward image editing application: by editing attributes, objects, or relationships in the scene graph, corresponding images can be generated. Fig. \ref{fig: edit_img} shows the results of this approach.

\section{Limitation}
We summarize statistics on the types of objects annotated in the LAION-Aesthetics~\cite{LAION_5B} and LAION-SG datasets. Among 10,000 samples, LAION-Aesthetics contains 12,263 distinct object types, which reduces to 5,811 after excluding proper nouns. In comparison, LAION-SG includes 1,429 types, all of which are common words without any proper nouns. This difference reflects a limitation of LAION-SG, as its vocabulary distribution is relatively less extensive. Furthermore, since LAION-SG focuses on scene graph that describe specific content within images, it is less sensitive to abstract cues such as historical context or stylistic elements. Integrating these control factors into the scene graph-to-image process remains a promising direction for future research.

Compared to manual annotation workflows, automated method significantly improves efficiency and reduces costs. However, it slightly lacks the precision achievable through human annotation, as discussed in \cref{subsec: Hallucinations of GPT-4o} and \cref{subsec: Failure Examples}. This limitation, however, is an inherent aspect of automated processes.


\section{Social Impact}
Scene graph to image generation holds great potential to benefit diverse fields, from content creation and education to virtual reality and simulation. By enabling the generation of realistic images from structured descriptions, this technology democratizes creative processes, allowing individuals with limited artistic skills to visualize complex ideas efficiently. Moreover, it can facilitate accessibility for users with disabilities, providing new ways to interact with visual content.

However, the technology also poses challenges, such as potential misuse for generating misleading, harmful content and negative bias. To mitigate these risks, the dataset and methods proposed in this work prioritize ethical considerations, including content moderation and bias reduction. Future research and collaboration across disciplines are essential to ensure that such technologies align with societal values while maximizing their positive impact.